\def\year{2022}\relax
\documentclass[letterpaper]{article} 
\usepackage{aaai22}  
\usepackage{times}  
\usepackage{helvet}  
\usepackage{courier}  
\usepackage[hyphens]{url}  
\usepackage{graphicx} 
\usepackage{natbib}  
\usepackage{caption} 
\DeclareCaptionStyle{ruled}{labelfont=normalfont,labelsep=colon,strut=off} 
\usepackage{algorithm}
\usepackage{algorithmic}

\usepackage{graphicx}
\usepackage{enumerate}
\usepackage{amsfonts}
\usepackage{graphics} 
\usepackage{multirow}
\usepackage{array}
\usepackage{calrsfs}
\usepackage{anyfontsize}
\usepackage{color}
\usepackage{amsmath}

\usepackage{newfloat}
\usepackage{listings}
\floatstyle{ruled}
\newfloat{listing}{tb}{lst}{}
\floatname{listing}{Listing}
\title{Explainable and Interpretable Diabetic Retinopathy Classification Based on Neural-Symbolic Learning}

\author{
    Se-In Jang\textsuperscript{\rm 1,2}, 
    Michaël J.A. Girard\textsuperscript{\rm 3,4,5} and 
    Alexandre H. Thiery\textsuperscript{\rm 6}
}
\affiliations{
    \textsuperscript{\rm 1}Gordon Center for Medical Imaging, Massachusetts General Hospital and Harvard Medical School, Boston, USA\\
    \textsuperscript{\rm 2}Center for Advanced Medical Computing and Analysis, Massachusetts General Hospital and Harvard Medical School, Boston, MA, USA\\
    \textsuperscript{\rm 3}Ophthalmic Engineering and Innovation Laboratory, Singapore Eye Research Institute, Singapore\\
    \textsuperscript{\rm 4}Duke-NUS Medical School, Singapore\\
    \textsuperscript{\rm 5}Institute for Molecular and Clinical Ophthalmology, Basel, Switzerland\\
    \textsuperscript{\rm 6}Department of Statistics and Data Science, National University of Singapore, Singapore\\

    sjang7@mgh.harvard.edu\textsuperscript{\rm 1}, mgirard@ophthalmic.engineering\textsuperscript{\rm 2,3,4}, a.h.thiery@nus.edu.sg\textsuperscript{\rm 5}

}

\begin{document}

\maketitle

\begin{abstract}
In this paper, we propose an explainable and interpretable diabetic retinopathy (ExplainDR) classification model based on neural-symbolic learning. 
To gain explainability, a high-level symbolic representation should be considered in decision making.
Specifically, we introduce a human-readable symbolic representation, which follows a taxonomy style of diabetic retinopathy characteristics related to eye health conditions to achieve explainability. 
We then include human-readable features obtained from the symbolic representation in the disease prediction.
Experimental results on a diabetic retinopathy classification dataset show that our proposed ExplainDR method exhibits promising performance when compared to that from state-of-the-art methods applied to the IDRiD dataset, while also providing interpretability and explainability.
\end{abstract}

\section{Introduction}
Diabetic Retinopathy (DR) is one of the leading causes of vision loss affecting the working age population worldwide \cite{garg2009diabetic}. 
Thanks to the success of deep learning, convolutional neural networks (CNNs) based deep learning approaches have been recently applied to DR classification problems \cite{schmidt2018artificial,ting2019artificial,ting2019deep}.
Most of the research efforts devoted to CNN-based DR classification methods have been devoted to designing robust neural architectures (e.g. ResNet and DenseNet) for enhanced classification accuracy \cite{pratt2016convolutional,yang2017lesion}.
Although deep-learning-based DR classification approaches have demonstrated excellent performance, understanding the decision making process remains a challenge because of the black-box nature of the deep learning methods. This lack of explainability has hindered the adoption of deep-learning based methods in clinical settings.


To gain confidence that developed deep learning methods are robust, 
researchers have designed and used visually interpretable tools. For instance, gradient-weighted class activation mapping (Grad-CAM) \cite{selvaraju2017grad} is a popular approach that can highlight suspected lesions  \cite{chetoui2020explainable}. However, most of these post-processing tools generate images (e.g. attention maps) that can only be interpreted by expert ophthalmologists. To circumvent this issue, in \cite{lalonde2020encoding}, a capsule network \cite{sabour2017dynamic} was adopted to encode visually interpretable feature scores for X-ray images in a human-level representation -- importantly, these scores can also be interpreted by radiologists. However, this approach could not be considered an explainable model per se since a taxonomy style of characteristics or attributes (such as eyes, a nose, and a mouth that can be used to define a given face) was not involved in the decision making process \cite{gunning2017explainable}.




In order to achieve interpretability and completeness for an explainable DR classification model, 
we have to understand how DR severity is defined clinically.
Table \ref{tbl.grading} summarizes grading criteria for DR severity.
Clinically, DR is diagnosed based on the presence of one or more
retinal lesions such as Microaneurysms (MA),
Hemorrhages (HE),
Soft Exudates (SE) and
Hard Exudates (EX)  \cite{yau2012global}.
In addition, Diabetic Macular Edema (DME) severity is also assessed based on the presence of EXs
in the macula region \cite{decenciere2014feedback}.
\begin{table}[t]
	\centering \tiny 
	\caption{Diabetic Retinopathy Severity Grading Criteria \cite{porwal2020idrid}. }\label{tbl.grading}  \vspace{-0.3cm}
	\begin{tabular}{|l|l|}
		\hline
		Severity  Grade &  Description \\
		\hline
		No DR: & No visible sign of abnormalities \\
		Mild NPDR*: & Presence of MAs only \\
		Moderate NPDR: & More than just MAs but less than severe NPDR \\
		\begin{tabular}[l]{@{}l@{}} Severe NPDR:  \\ \\ \end{tabular} & \begin{tabular}[l]{@{}l@{}}   $>20$ intraretinal HEs, Venous beading, \\ Intraretinal microvascular abnormalities, No signs of PDR   \end{tabular}  \\
		\begin{tabular}[l]{@{}l@{}} PDR**:    \\  \end{tabular} & \begin{tabular}[l]{@{}l@{}}  Neovascularization, Vitreous/pre-retinal HE   \end{tabular}   \\
		\hline 
	\end{tabular} \\
	{\scriptsize *NPDR: Non-Proliferative DR, **PDR: Proliferative DR} 
	\vspace{-0.5cm}
\end{table}


Neural-symbolic learning \cite{garcez2015neural, garcez2020neurosymbolic} is a suitable approach to produce computational tools for integrated machine learning and reasoning for explainability \cite{besold2017neural}. 
Neural-symbolic learning uses deep neural networks to generate high-level symbolic representation that humans can understand.
Logical operations are then conducted using symbolic representation for decision making.
In \cite{yi2018neural}, a neural-symbolic learning system for visual question answering was presented to find an answer from a structural scene representation.
This system encoded an image into a compact symbolic representation and then performed
symbolic program execution that included logical operations manually designed for reasoning.
However, due to the manual design, updating logics for improving performances is not an easy task since the logics should consider relationships between each other.

In this paper, we propose an explainable and interpretable diabetic retinopathy (ExplainDR) classification model based on neural-symbolic learning to generate a human-readable symbolic representation. 
The proposed symbolic representation follows a taxonomy style of diabetic retinopathy characteristics consisting of several abnormalities such as MA, HE, SE and EX via a deep neural network for segmentation.
The proposed human-readable feature representation is meant to be directly interpretable by both ophthalmologists and patients. 


In this paper, we aim to develop a neural-symbolic AI approach to accurately diagnose DR. Such an approach may be of clinical value, because we first generate high-level symbolic representations that are subsequently used to make a DR diagnosis. In other words, our approach has the advantage to remain easily interpretable by both clinicians and patients. The algorithm was tested on the the IDRiD dataset \cite{porwal2020idrid}, and heavily relied on lesion segmentation and disease severity gradings.


\section{Related Works}
\subsection{Visually interpretable based deep learning models}
In order to improve the black box based deep learning models,
visually interpretable tools \cite{zhou2016learning, lundberg2017unified, sundararajan2017axiomatic, smilkov2017smoothgrad} for map generation (e.g. attention maps) have been recently applied to DR problems.
In \cite{wang2017zoom}, an attention network was used as a clustering method to generate an attention map that can highlight the suspected lesions. This can also be achieved with
Class Activation Mapping (CAM) \cite{zhou2016learning, jiang2019interpretable}.
In \cite{wang2017diabetic}, a regression based activation map was developed to include severity level information in the generated saliency map.
In \cite{chetoui2020explainable}, a Grad-CAM method that can evaluate the suspected lesions without requiring architectural changes or re-training \cite{selvaraju2017grad}, was adopted to use different CNN architectures for improving visual interpretability.
In \cite{linellg2020}, a combination of lower-layer and higher-layer saliency maps was developed to accurately locate the lesions. 
Although the above methods could provide clinical value, they still could not explain \textit{why and how} the developed models could visually localize the suspected lesions.

\subsection{Neural-Symbolic Learning}
The goal of neural-symbolic learning is to provide a coherent, unifying view for logic and connectionism to contribute to the modelling and understanding of cognition and, thereby, behavior \cite{garcez2015neural}.
The neural-symbolic learning includes a neural network implementation of a logic, a logical characterisation of a neural network system and  a hybrid learning system that profitably achieves  symbolic and connectionist approaches together to artificial intelligence.
Deep neural networks can learn complex input data such as images, audio and text to generate high-level representations, which are useful in decision making \cite{garcez2019neural}. 
A logic network on top of a deep neural network to learn the relations of those abstractions, can then help systems to be able to explain itself.
In \cite{manhaeve2018deepproblog}, DeepProbLog was developed by combining an end-to-end learning with reasoning, where outputs of the neural networks were applied as inputs to ProbLog \cite{de2007problog}.
In \cite{riegel2020logical}, a neural-symbolic framework called logical neural networks (LNN) was designed to simultaneously provide key properties of both neural networks for learning and symbolic logics for knowledge and reasoning.
LNN considers every neuron to have a meaning as a component of a formula in a weighted real-valued logic.
In LNN, an idea of a 1-to-1 correspondence between neurons and the elements of logical formulae was presented by observing the weights of neurons that can act like AND or OR operations.
Based on this idea, LNN has achieved a differentiable model that can minimize a logical loss function for refutation of logical contradiction.

\section{Explainable and Interpretable Diabetic Retinopathy Classification}
\begin{figure}[t]
	\centering
	\includegraphics[width=0.45\textwidth]{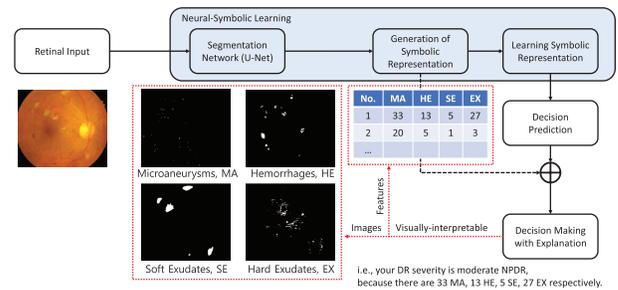}  \vspace{-0.3cm}
	\caption{An overview of the proposed explainable and interpretable diabetic retinopathy classification.} \label{fig.ProposedFigure} 	\vspace{-0.5cm}
\end{figure}
In this section, we propose an explainable and interpretable diabetic retinopathy (ExplainDR) classification method based on neural-symbolic learning.
Fig.~\ref{fig.ProposedFigure} illustrates an overview of the proposed ExplainDR method.
Our proposed neural-symbolic learning method includes a U-Net segmentation network \cite{ronneberger2015u} used to generate a high-level symbolic representation and a fully connected network (FCN) for learning the generated symbolic representation to predict decision instead of designing logical operations \cite{towell1994knowledge}.
The U-Net segmentation network extracts a higher-level representation in a symbolic space than the pixel-level representation.
To produce the high-level symbolic representation in a taxonomy style, we train the U-Net segmentation network using four segmentation labels, namely Microaneurysms (MA),
Hemorrhages (HE),
Soft Exudates (SE) and
Hard Exudates (EX) which are the main factors to decide about DR severity.
Based on the four output images $I^i$, $1 \le i \le 4$ produced by the segmentation network for each eye condition  (i.e. $i=1$ for MA), we extract a human-readable feature vector as symbolic representation using a quantization technique. This feature vector counts the segmented regions in each segmentation output image $I^i$ by setting
\begin{equation}
	S_{}^i = \left\{ {x_{}^j} \right\}_{j = 1}^{N^i}
\end{equation}
where $S_{}^i$ is a set of the segmented regions $x^j$ in $I^i$ and  
${N^i}$ is the number of segmented regions within each set $S_{}^i$. 
The human-readable feature vector is then given by
\begin{equation}
	F_{sym} = \big[ \, {\left| {S_{}^1} \right|,\left| {S_{}^2} \right|,\left| {S_{}^3} \right|,\left| {S_{}^4} \right|}  \, \big] \in \mathbb{N}^4,
\end{equation}
where $\left| {S_{}^i} \right|$ is the number of segmented regions in $I^i$.
The human-readable feature vector is trained using the FCN instead of performing the logical operations to avoid the efforts of designing considerable logic combinations for decision making.

For instance, from an unseen test image, the human-readable feature vector is obtained from each segmented output through the trained segmentation network.
Based on the trained FCN, the decision prediction is performed using the human-readable feature vector.
We then generate explanation by combining the human-readable feature vector and the predicted decision as follows:
\begin{itemize} 
	\item The DR diagnosis of ``image 1'' is ``moderate NPDR'' because there are 33 MA, 13 HE, 5 SE and 27 EX regions, respectively.
	\item The DR diagnosis of ``image 2'' is ``mild NPDR'' because there are 20 MA, 5 HE, 1 SE and 3 EX regions, respectively.
\end{itemize}
Additionally, similar to other interpretable DR methods, the visually interpretable images (i.e. segmented images) are also provided.
Therefore, we achieve an explainable and interpretable DR classification method, which includes human-readable symbolic representation in the decision making process, whereas typical AI black-box models only address pixel-level representations.


\subsection{Extension of the symbolic representation}
Our proposed human-readable feature vector consists of the simple symbolic representation in only four dimensions, and for the four eye conditions (e.g. MA, HE, SE and EX).
In order to improve the simple symbolic representation,
we propose to consider the sizes of the segmented lesions for better symbolic representation while removing false or noisy segmented lesions.
Each segmented lesion ${x^j}$ is classified into one of three subsets: small, medium or large size as follows:
\begin{equation} \label{eq.ext}  
	\scriptsize 
	\begin{aligned}
		S_{s}^{i} &= \left\{ {{x^j}: \tau _0^{} < {s^j} \le \tau _1^{}{\rm{, }}\forall j} \right\},\\
		S_{m}^{i} &= \left\{ {{x^j}:\tau _1^{} < {s^j} \le \tau _2^{}{\rm{, }}\forall j} \right\},\\
		S_{l}^{i} &= \left\{ {{x^j}:\tau _2^{} < {s^j} \le \tau _3^{}{\rm{, }}\forall j} \right\},
	\end{aligned}
\end{equation}
where the size $s^j$ is given by the number of the connected pixels in each segmented lesion $x^j$.
$\tau$ is a threshold that experimentally defines the small, medium and large sizes of the segmented lesions.
The improved human-readable feature vector is then given by:
\begin{equation} 
	\footnotesize
	\begin{aligned}
		F_{sml} = \left[ {\left| {S_{ s}^1} \right|,\left| {S_{m}^1} \right|,\left| {S_{l}^1} \right|, \ldots ,\left| {S_{s}^4} \right|,\left| {S_{m}^4} \right|,\left| {S_{l}^4} \right|} \right] \in \mathbb{N}^{12}.
	\end{aligned}
\end{equation}
We note that the extended human-readable feature vector is still under a taxonomy style that can offer logical explanation according to the different sizes of the segmented lesion within each eye condition.

\section{Experiments}
\subsection{Experimental settings}
In our experiment, 
we use the Indian Diabetic Retinopathy Image Dataset (IDRiD)\footnote{https://idrid.grand-challenge.org} \cite{porwal2020idrid},
since this is the one public dataset that provides both lesion segmentation and disease severity gradings.
The images have the resolution of $4288 \times 2848$ pixels.
Each image is resized to $1024 \times 1024$ pixels.
In the lesion segmentation dataset, four labels such as 
Microaneurysms (MA),
Hemorrhages (HE),
Soft Exudates (SE) and
Hard Exudates (EX) are included.
In the severity grading dataset, five labels for diabetic retinopathy (DR) such as
no DR, mild NPDR, moderate NPDR, severe NPDR and PDR are provided.
Additionally, three labels for diabetic macular edema (DME) such as 
no EX, presence of EX outside and within the macula center
are also given.
The lesion segmentation dataset has 187 training images and 95 test images in total 282 images.
The severity grading dataset provides 413 training images and 103 test images in total 516 images.

In the IDRiD challenge \cite{porwal2020idrid}, they provided a specific accuracy evaluation metric counts when the following condition is satisfied: 
\begin{equation} \footnotesize 
	\left( {y_{DR}^{} =  = \hat y_{DR}^{}} \right) \quad {\rm{and}} \quad \left( {y_{DME}^{} =  = \hat y_{DME}^{}} \right), 
\end{equation}
where $y$ is a true label, and $\hat y$ is a predicted label for DR and DME.
In Equation \eqref{eq.ext}, the thresholds are experimentally set at $\tau _0^{}=10$, $\tau _1^{}=500$, $\tau _2^{}=1000$ and $\tau _3^{}=10000$ respectively.

In the segmentation network, the ResNet34 structure \cite{he2016deep} is used with the Adam optimizer following a batch size of 2, a learning rate of 0.0001 and a dropout probability of 0.1 for 20 epochs with early stopping. 
The data augmentation of the segmentation networks includes random flipping, gamma contrast with a range (0.5, 1.5) and a contrast limited adaptive histogram equalization.
The FCN layers are given by: [12, 25, 50, 75, 100, 75, 50, 25, 12].
In the FCN layers, the Adam optimizer is adopted with a batch size of 16, a learning rate of 0.01 and a dropout probability of 0.1 for 20 epochs with early stopping.
The segmentation network is first trained using the lesion segmentation training set.
The FCN layers are then trained using the proposed symbolic feature vectors obtained from the severity grading training set via the trained segmentation network.
We split the training sets into $80\%$ for training and $20\%$ for validation.
\begin{figure}[t]\centering
	\includegraphics[width=0.38\textwidth]{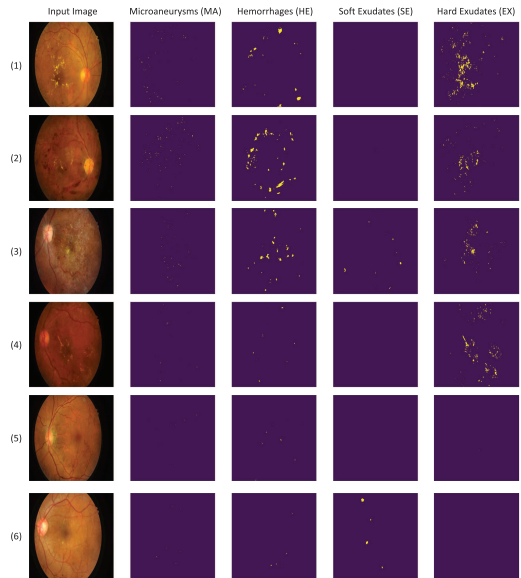} \vspace{-0.3cm}
	\caption{Segmentation results of the proposed ExplainDR in the severity grading dataset.} \label{fig.SegFigure}  \vspace{-0.3cm}
\end{figure}

\subsection{Results}
In order to observe the effect of our proposed ExplainDR method, 
we conduct an ablation study to evaluate the extension of the human-readable feature vector.
We compare the proposed ExplainDR method with the state-of-the-art methods using the IDRiD dataset.
Fig. \ref{fig.SegFigure} qualitatively shows the segmentation results for eye conditions such as MA, HE, SE and EX using six images from the severity grading dataset.
According to small, medium and large (sml) size regions of each eye condition, the six extracted human-readable feature vectors for each image are as follows:
\begin{enumerate}[(1)]
	\item smlMA: 37,   0,  0, smlHE: 26,   2, 2, smlSE: \phantom{0}0,  0,   0, smlEX: 197,  5,   3
	\item smlMA: 59,  0,  0, smlHE: 54,  4,  4, smlSE: \phantom{0}0,  0,   0, smlEX: \phantom{0}96,  2,  0
	\item ...
	\item smlMA: \phantom{0}1, 0, 0, smlHE: \phantom{0}5, 0, 0, smlSE: \phantom{0}2, 1, 1, smlEX: \phantom{00}0, 0, 0
\end{enumerate}
The explanation along with the predicted decision using the human-readable features are  generated as follows:
\begin{enumerate}[(1)]
	\item The image 1 is classified as severe NPDR  because 37 small MAs, 26 small HEs, 2 medium HEs, 2 large HEs, 197 small EXs, 5 medium EXs and 3 large EXs are detected.
	\item The image 2 is classified as PDR   because 59 small MAs, 54 small HEs, 4 medium HEs, 4 large HEs, 96 small EXs, and 2 medium EXs   are detected.
	\item ... 
	\item The image 6 is classified as mild NPDR  because 1 small MA, 5 small HEs, 2 small HEs, 1 medium HE, 1 large HE are detected.
\end{enumerate}
Here, we note that the above explanations can be compared to  the severity grading criteria shown in Table \ref{tbl.grading} by summing all the numbers of the small, medium and large size regions for each eye condition. This helps non-experts to analyze the generated explanations for self-diagnosis.

To observe the impact of symbolic feature extension of the proposed ExplainDR method, 
Table \ref{tbl.aba} shows an ablation study for:
1) ExplainDR with 4 dimensions  of the simple symbolic features and
2) ExplainDR with 12 dimensions  of the extended symbolic features.
The extension of the symbolic representation outperforms that of the simple symbolic representation since the detailed categorization of the simple symbolic representation provides more discriminative symbolic representation than the simple symbolic representation.
For performance comparison, 
Table \ref{tbl.results} summarizes accuracy performances of the proposed ExplainDR method and the state-of-the-art methods \cite{porwal2020idrid}.
The proposed method without utilizing any external dataset (e.g. Kaggle\footnote{https://www.kaggle.com/c/diabetic-retinopathy-detection}, Messidor\footnote{https://www.adcis.net/en/third-party/messidor} and DiaretDB1\footnote{http://www2.it.lut.fi/project/imageret/diaretdb1}) shows the second-best performance with interpretable images and texts in the leaderboard on the IDRiD dataset. Whereas, the state-of-the-art methods with external datasets provide the accuracy performances without any explanation.
\begin{table}[t!] 
	\centering \tiny
	\caption{An ablation study of the proposed ExplainDR method.}\label{tbl.aba} \vspace{-0.3cm}
	\begin{tabular}{|@{\hskip0.5pt}l@{\hskip0.5pt}|@{\hskip0.5pt}c@{\hskip0.5pt}|}
		\hline
		Name                      & Accuracy  \\ \hline
		ExplainDR + Simple Symbols    & 0.4757            \\
		ExplainDR + Extended Symbols  & 0.6019                  \\
		\hline                                  
	\end{tabular}
\end{table}
\begin{table}[t!] 
	\centering \tiny
	\caption{The leaderboard on the DR and DME test sets in the IDRiD challenge.}\label{tbl.results}  \vspace{-0.3cm}
	\begin{tabular}{|@{\hskip0.5pt}l@{\hskip0.5pt}|@{\hskip0.5pt}c@{\hskip0.5pt}|@{\hskip0.5pt}l@{\hskip0.5pt}|@{\hskip0.5pt}c@{\hskip0.5pt}|@{\hskip0.5pt}l@{\hskip0.5pt}|} 
		\hline
		Name                        & Accuracy & Approach                                    & Input Size & \begin{tabular}[c]{@{}l@{}}External \\ Dataset\end{tabular} \\
		\hline
		LzyUNCC                     & 0.6311   & ResNet + Deep Aggregation                                  & 896 $\times$ 896   & Kaggle            \\
		ExplainDR    				& 0.6019   & Symbols  + FCN & 1024 $\times$ 1024    &  -        \\
		VRT                         & 0.5534   & CNN                                         & 640 $\times$ 640    & Kaggle, Messidor  \\
		Mammoth                     & 0.5146   & DenseNet                                    & 512 $\times$ 512    & Kaggle            \\
		HarangiM1                   & 0.4757   & AlexNet + GoogLeNet                           & 224 $\times$ 224    & Kaggle            \\
		AVSASVA                     & 0.4757   & ResNet + DenseNet                             & 224 $\times$ 224    & DiaretDB1         \\
		HarangiM2                   & 0.4078   & AlexNet + Handcrafted                 & 224 $\times$ 224    & Kaggle     
		\\
		\hline      
	\end{tabular}
	\vspace{-0.5cm}
\end{table}
\section{Conclusion}
This paper presented an explainable and interpretable diabetic retinopathy (ExplainDR) classification method based on neural-symbolic learning which generated a high-level symbolic representation via a segmentation network. 
The generated symbolic representation was extended according to the sizes of the segmented lesions to produce more discriminative symbolic representation.
The DR severity is predicted by the fully connected network, which was trained using the extended  symbolic representation.
We qualitatively showed that our proposed symbolic representation was human-readable in the taxonomy style associated with the eye health conditions, as well as an explanation with the reasons of the DR severity.
The proposed ExplainDR method showed promising performances to the state-of-the-art methods
in terms of classification accuracies on the IDRiD dataset as well as providing interpretability and explainability.

The limitations of our works are: 
1) The accuracy and explainablity performances of the proposed ExplainDR are affected by the quality of the segmentation results;
2) Different decision outputs can be observed due to the nature of stochastic learning (e.g. FCN); and
3) An enhanced design is needed to adopt other datasets if there is no annotation of the lesion segmentation and the DR classification together. 
Our future works accordingly are as follows:
1) Study of the effect of the segmentation performance;
2) Use of least-squares based methods as a deterministic learning approach instead of the stochastic learning approach; and
3) Study of adoption of other datasets without annotation of the lesion segmentation.


\bibliography{references.bib}

\end{document}